\documentclass[11pt]{article}

\usepackage{acl}

\usepackage{times}
\usepackage{latexsym}

\usepackage[T1]{fontenc}
\usepackage[utf8]{inputenc}

\usepackage{microtype}
\usepackage{inconsolata}

\usepackage{graphicx}
\usepackage{booktabs}
\usepackage{multirow}
\usepackage{amsmath}
\usepackage{amssymb}

\usepackage{xcolor}
\usepackage{pdfpages} 

\title{EvasionBench: A Large-Scale Benchmark for Detecting \\Managerial Evasion in Earnings Call Q\&A}

\author{
  Shijian MA$^{1,\dagger}$, Yan LIN$^{2,\dagger,*}$, Yi YANG$^{1}$ \\
  $^{1}$Hong Kong University of Science and Technology \\
  $^{2}$University of Macau \\
  \texttt{mas8069@foxmail.com, yanlin@um.edu.mo, imyiyang@ust.hk} \\
  $^{\dagger}$Equal contribution \quad $^{*}$Corresponding author
}

\begin{document}
\maketitle

\begin{abstract}
We present \textbf{EvasionBench}, a comprehensive benchmark for detecting evasive responses in corporate earnings call question-and-answer sessions.
Drawing from 22.7 million Q\&A pairs extracted from S\&P Capital IQ transcripts, we construct a rigorously filtered dataset and introduce a three-level evasion taxonomy: \textit{direct}, \textit{intermediate}, and \textit{fully evasive}.
Our annotation pipeline employs a Multi-Model Consensus (MMC) framework, combining dual frontier LLM annotation with a three-judge majority voting mechanism for ambiguous cases, achieving a Cohen's Kappa of 0.835 on human inter-annotator agreement.
We release: (1) a balanced 84K training set, (2) a 1K gold-standard evaluation set with expert human labels, and (3) \textbf{Eva-4B}, a 4-billion parameter classifier fine-tuned from Qwen3-4B that achieves 84.9\% Macro-F1, outperforming Claude Opus 4.5, GPT-5.2, and Gemini 3 Flash.
Our ablation studies demonstrate the effectiveness of multi-model consensus labeling over single-model annotation.
EvasionBench fills a critical gap in financial NLP by providing the first large-scale benchmark specifically targeting managerial communication evasion.
\end{abstract}

\section{Introduction}
\label{sec:intro}

Evasive communication---responding to questions without truly answering them---is a pervasive phenomenon across human discourse.
Politicians deflect uncomfortable questions in interviews \citep{bull1993how,clayman2001answers}, witnesses hedge in legal depositions \citep{bachenko2008verification}, and corporate executives sidestep analyst inquiries during earnings calls \citep{hollander2010does,gow2021non}.
Detecting such evasion has concrete implications: in politics, it undermines democratic accountability; in legal contexts, it obscures truth-finding; and in financial markets, evasive management communication predicts subsequent earnings misses and stock underperformance \citep{barcellos2025overcoming,larcker2012detecting}.

Despite evasion's ubiquity, it remains underexplored in NLP.
Unlike sentiment analysis---which has canonical benchmarks like SST and Financial PhraseBank---or question answering with SQuAD, no large-scale benchmark exists for evasion detection.
This gap stems from two challenges: (1) the inherently subjective nature of ``evasiveness,'' which lies on a spectrum from direct to fully evasive, and (2) the prohibitive cost of expert annotation at scale.
While sentiment captures \textit{what} is said, evasion detection addresses \textit{whether} the question was actually answered---a fundamentally different, discourse-level phenomenon rooted in Gricean pragmatics \citep{grice1975logic}.

Corporate earnings calls offer an ideal testbed for evasion research.
They provide: (a) high-stakes, adversarial Q\&A with clearly defined speaker roles (analysts vs.\ executives), (b) abundant publicly available transcripts spanning decades, and (c) well-defined question types (quantitative, temporal, binary, causal) that enable operationalization of ``directness.''
Prior work has examined non-answers in this domain \citep{gow2021non,bamber2025can}, but datasets remain limited in scale.

In this paper, we introduce \textbf{EvasionBench}, the first large-scale benchmark for evasion detection, using earnings calls as our primary domain.
While our data is domain-specific, we hypothesize that both our annotation framework and taxonomy could transfer to other adversarial Q\&A settings (e.g., political interviews, legal depositions), though cross-domain validation remains future work.
Our contributions are:

\begin{itemize}
    \item \textbf{EvasionBench}, a large-scale dataset comprising 84K balanced training samples and 1K human-validated evaluation samples, derived from 22.7M raw Q\&A pairs spanning 2002--2022.

    \item A \textbf{Multi-Model Consensus (MMC) framework} for scalable annotation that combines dual frontier LLM labeling with three-judge arbitration, achieving Cohen's Kappa of 0.835 against human annotators.

    \item \textbf{Eva-4B}, a 4-billion parameter classifier that achieves 84.9\% Macro-F1, outperforming Claude Opus 4.5, GPT-5.2, and Gemini 3 Flash on this task.

    \item Comprehensive analysis demonstrating that multi-model consensus labeling significantly outperforms single-model annotation, with ablations showing +4.3 pp Macro-F1 improvement.
\end{itemize}

\section{Related Work}
\label{sec:related}

\subsection{Evasion Across Domains}

Evasive communication has been studied extensively across multiple fields, though NLP benchmarks remain scarce.
In political communication, \citet{bull1993how} identified 43 distinct techniques politicians use to avoid answering questions, while \citet{clayman2001answers} distinguished between overt evasion (explicit refusal) and covert evasion (appearing to answer while sidestepping the core).
\citet{bavelas1990equivocal} theorized that equivocation arises from avoidance-avoidance conflicts where all direct answers carry costs---a framework directly applicable to corporate executives facing analyst scrutiny.

In legal and forensic contexts, researchers have developed computational approaches to detect deceptive language in testimony \citep{bachenko2008verification,vrij2008detecting}.
\citet{newman2003lying} identified linguistic markers of deception using LIWC, achieving 61--67\% classification accuracy.
Crucially, evasion differs from deception: while deception involves false statements, evasion represents strategic non-responsiveness where speakers appear cooperative but avoid the question core \citep{perezrosas2015deception}.

From a pragmatic perspective, evasion can be analyzed as a violation of Grice's maxim of Relation \citep{grice1975logic}, often realized through hedging devices \citep{lakoff1973hedges,hyland1998hedging} and indirect speech acts \citep{searle1975indirect}.
\citet{brown1987politeness} connected such indirectness to face-threatening act mitigation, explaining why speakers strategically employ vague language.

Despite this rich theoretical foundation across disciplines, NLP lacks a canonical benchmark for evasion detection---unlike sentiment \citep{socher2013sst}, stance \citep{mohammad2016semeval}, or factual QA \citep{rajpurkar2016squad}.
Our work addresses this gap by providing the first large-scale, systematically annotated evasion benchmark.

\subsection{Evasion in Financial Communication}

Research on managerial evasion during earnings calls has grown substantially in accounting and finance.
\citet{hollander2010does} provide empirical evidence that managers strategically choose what to disclose, with silence interpreted negatively by investors.
\citet{gow2021non} find that approximately 11\% of analyst questions receive non-answers, with product-related questions frequently evaded in competitive environments.
\citet{bamber2025can} offer theoretical frameworks for understanding why managers evade questions, while \citet{barcellos2025overcoming} examine how investor suspicion interacts with evasive communication.
\citet{nuaimi2025detecting} propose a psychological taxonomy for evasive answers; however, their dataset relies solely on single-model LLM annotation without human validation, raising concerns about label reliability and potential model-specific biases.

\subsection{Financial NLP}

Financial text analysis has evolved from lexicon-based methods \citep{loughran2011liability} to neural approaches.
\citet{araci2019finbert} introduced FinBERT for financial sentiment analysis, demonstrating the value of domain-specific pretraining.
\citet{koval2023forecasting} predict earnings surprises from conference call transcripts using LLMs.
\citet{huang2014tone} show that managers strategically manage linguistic tone.
However, these works focus on sentiment or prediction rather than discourse-level phenomena like evasion, which requires assessing whether a response actually addresses the question asked.

\subsection{Scalable Annotation with LLMs}

The prohibitive cost of expert annotation has driven interest in LLM-based labeling approaches.
Using LLMs as evaluators has gained traction for scalable assessment \citep{zheng2023judging}, though \citet{chen2024humans} identify systematic biases including position bias and verbosity preference.

Several methodologies address annotation quality and consistency.
Self-consistency \citep{wang2022self} samples multiple reasoning paths from a single model.
RLAIF \citep{lee2023rlaif} uses AI feedback to scale preference learning.
Tri-training \citep{zhou2005tritraining} leverages three classifiers to label unlabeled data through agreement.
Recent surveys on LLM-based annotation \citep{tan2024llm} validate consensus-based approaches for improving label quality.

Our MMC framework draws on these insights, combining multiple \textit{heterogeneous} frontier LLMs (rather than homogeneous classifiers) with randomized presentation order to mitigate individual model biases.
This design choice is motivated by our finding that different LLMs exhibit distinct labeling tendencies (Section~\ref{sec:annotation}).

\subsection{Summary and Positioning}

Table~\ref{tab:related_comparison} synthesizes prior work on evasion detection in financial communication. Three gaps emerge: (1) existing datasets lack scale---the largest prior corpus contains fewer than 15K samples with limited human validation; (2) annotation often relies on single-model LLM labeling without consensus mechanisms, introducing systematic bias; (3) no prior work provides both large-scale training data and rigorously human-validated evaluation sets. EvasionBench addresses all three limitations.

\begin{table}[t]
\centering
\small
\setlength{\tabcolsep}{3pt}
\begin{tabular}{lccccc}
\toprule
\textbf{Work} & \textbf{Scale} & \textbf{Labels} & \textbf{Human} & \textbf{Multi-M} \\
\midrule
\citet{gow2021non} & 2.1K & 2 & \checkmark & -- \\
\citet{nuaimi2025detecting} & 12K & 7 & -- & -- \\
\textbf{EvasionBench} & \textbf{85K} & 3 & \checkmark & \checkmark \\
\bottomrule
\end{tabular}
\caption{Comparison with prior evasion datasets. Labels = number of categories; Human = human validation; Multi-M = multi-model consensus annotation.}
\label{tab:related_comparison}
\end{table}

\section{Task Definition and Taxonomy}
\label{sec:task}

\subsection{Problem Formulation}

Given a question $q$ posed by a financial analyst and a response $a$ from corporate management, the task is to classify the response into one of three evasion levels: \textit{direct}, \textit{intermediate}, or \textit{fully evasive}.

\subsection{Evasion Taxonomy}

Prior taxonomies range from binary classifications (evasive vs.\ non-evasive; \citealp{gow2021non}) to fine-grained psychological categories (\citealp{nuaimi2025detecting} propose seven types including ``explicit refusal,'' ``deflection,'' and ``information flooding'').
We adopt a \textbf{three-level ordinal scale} for three reasons:
\begin{enumerate}
    \item \textbf{Empirical grounding}: Pilot annotation with five levels yielded low inter-annotator agreement ($\kappa < 0.5$); collapsing to three restored reliability ($\kappa = 0.83$).
    \item \textbf{Actionable granularity}: Binary classification loses the distinction between ``partial answer'' and ``complete deflection''---information critical for downstream applications like investor alerting.
    \item \textbf{Alignment with Gricean pragmatics}: Our levels map to degrees of Relation maxim violation \citep{grice1975logic}: full adherence (direct), partial violation (intermediate), and complete violation (fully evasive).
\end{enumerate}

Our taxonomy definitions, refined through iterative pilot annotation:

\paragraph{Direct} The response explicitly and completely addresses the question core. Typical features include specific figures, clear yes/no stances, or direct explanations.

\paragraph{Intermediate} The response provides related context but sidesteps the specific ask. Indicators include hedging language (``I think,'' ``it's possible''), conditional framing, or answering adjacent topics.

\paragraph{Fully Evasive} The question is ignored, explicitly refused, or the response is entirely off-topic. Examples include explicit refusal (``we don't provide that granularity''), information flooding, or silent pivoting.

\subsection{Illustrative Examples}

Table~\ref{tab:examples} presents representative examples from EvasionBench for each evasion category, illustrating the subtle distinctions our taxonomy captures.

\begin{table*}[t]
\centering
\small
\begin{tabular}{p{1.6cm}p{6.4cm}p{6.4cm}}
\toprule
\textbf{Label} & \textbf{Question} & \textbf{Answer} \\
\midrule
\textbf{Direct} &
\textit{``What exactly did you do with respect to those employees that dropped the turnover so much? Was it the wage increase?''} &
``No. So it was having good support for them, having coaches and supervisors... We've put recruiters in every district office... We've got better onboarding and training programs...'' \\
\midrule
\textbf{Interm.} &
\textit{``How are you thinking about your balance sheet and opportunities in the space given joint ventures, M\&A?''} &
``Yes, you're mentioning things that we internally think about a lot... we have a very strong balance sheet with positive net cash. So I think we're in a very strong position and we continue to evaluate opportunities...'' \\
\midrule
\textbf{Fully Ev.} &
\textit{``You mentioned an exciting pipeline of new products. Could you provide us with a sneak preview?''} &
``Well, I probably can't give you too much of a sneak preview. But all I can say is that historically, our business has been great at planning ahead... But you will have to see it when it comes out.'' \\
\bottomrule
\end{tabular}
\caption{Representative examples from EvasionBench. \textbf{Direct}: Manager explicitly lists specific actions. \textbf{Intermediate}: Response discusses balance sheet but avoids the M\&A question. \textbf{Fully Evasive}: Manager explicitly refuses to preview products.}
\label{tab:examples}
\end{table*}

\subsection{Question Core Types}

We identify four primary question core types:
\begin{itemize}
    \item \textbf{Quantitative}: Specific numbers, percentages, or magnitudes
    \item \textbf{Temporal}: Timelines, dates, or cadences
    \item \textbf{Binary}: Clear yes/no regarding plans or status
    \item \textbf{Causal/Directional}: Reasons for trends or future outlook
\end{itemize}

\section{Data Collection and Filtering}
\label{sec:data}

\subsection{Raw Data Source}

We source data from the S\&P Capital IQ Full Text Dataset, containing 22.7M Q\&A pairs from 1.38M earnings call transcripts spanning 420K unique speakers. After three-stage filtering, 11.27M high-quality pairs remain.

\subsection{Three-Stage Quality Filtering}

We apply a rigorous filtering pipeline:

\paragraph{Stage 1: Q\&A Pair Extraction}
Identify analyst questions (Type 3) and management answers (Type 4) with sequential component ordering. Remove operator instructions and pleasantries.

\paragraph{Stage 2: Quality Filtering}
Questions must contain ``?''; answers must exceed 30 characters; no transcription markers ([indiscernible], [ph], [inaudible]).

\paragraph{Stage 3: Substantial Content Selection}
Combined question and answer length $\geq$ 500 characters.

After filtering, 11.27M Q\&A pairs (49.6\%) remain.

\section{Multi-Model Consensus Annotation}
\label{sec:annotation}

\subsection{Framework Overview}

\begin{figure*}[t]
    \centering
    \includegraphics[width=\textwidth]{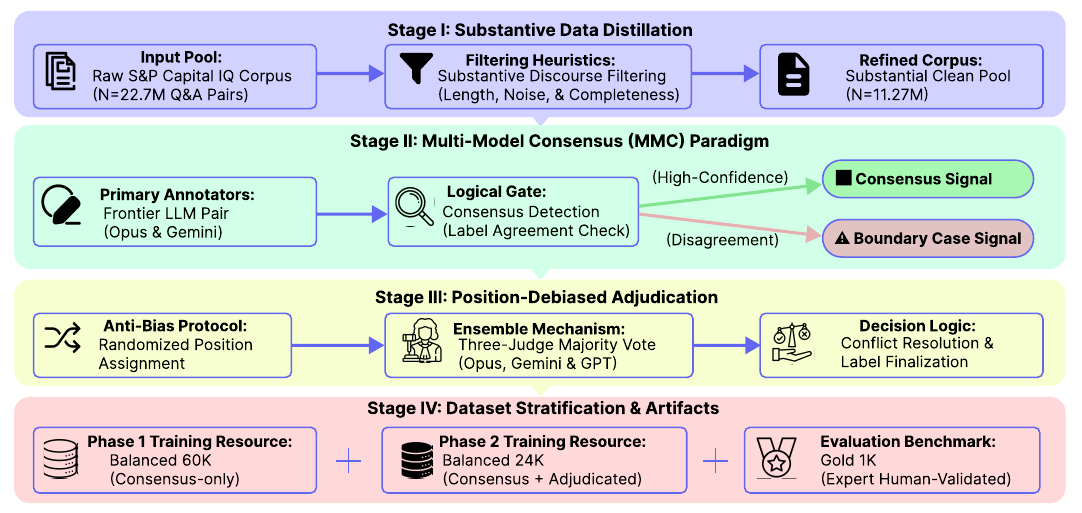}
    \caption{Multi-Model Consensus (MMC) annotation pipeline. Stage I: Data distillation from raw transcripts. Stage II: Dual LLM annotation with Claude Opus 4.5 and Gemini 3 Flash. Stage III: Three-judge arbitration for disagreement cases. Stage IV: Final balanced datasets.}
    \label{fig:mmc_pipeline}
\end{figure*}

Our Multi-Model Consensus (MMC) framework leverages agreement between frontier LLMs as a signal of annotation reliability. Figure~\ref{fig:mmc_pipeline} illustrates the complete pipeline.

\subsection{Stage I: Dual LLM Annotation}

We employ Claude Opus 4.5 and Gemini 3 Flash as primary annotators, each independently labeling samples with structured prompts.

\subsection{Stage II: Consensus Detection}

Samples where both models agree form the \textbf{consensus set}. Disagreement samples proceed to arbitration.

\subsection{Stage III: Three-Judge Arbitration}

For disagreement cases (3,645 samples, 16.1\% of total), three judge models (Claude Opus 4.5, Gemini 3 Flash, GPT-5.2) independently evaluate which original annotation is more accurate. Majority voting determines the final label.

\begin{figure}[t]
    \centering
    \includegraphics[width=\columnwidth]{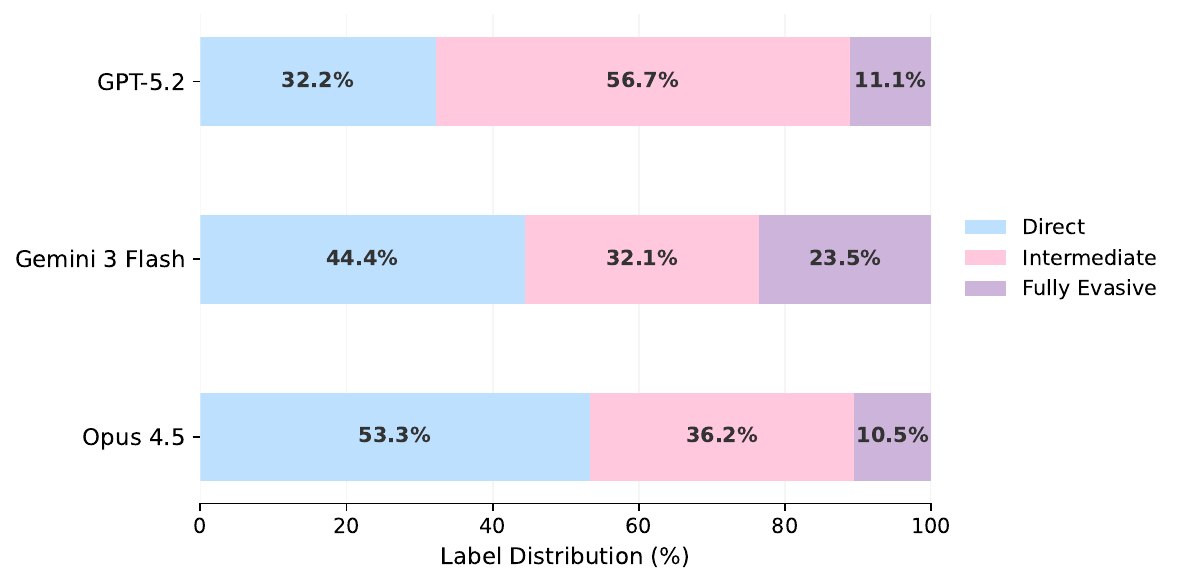}
    \caption{Label distribution across three judges on 3,645 disagreement samples. Each model exhibits distinct tendencies: Opus (conservative, 53\% direct), Gemini (strict, 24\% fully evasive), GPT-5.2 (balanced, 57\% intermediate).}
    \label{fig:judge_distribution}
\end{figure}

\paragraph{Why Multi-Model Consensus?}
Figure~\ref{fig:judge_distribution} reveals systematic differences in judge tendencies: Opus prefers \textit{direct} labels (53.3\%), Gemini assigns more \textit{fully\_evasive} (23.5\%), while GPT-5.2 favors \textit{intermediate} (56.7\%). This demonstrates that single-model annotation introduces systematic bias, motivating our multi-model consensus approach.

\paragraph{Anti-Bias Mechanism} To eliminate position bias (primacy effect), we randomize which model's prediction appears first in the judge prompt (seed=42 for reproducibility).

\subsection{Final Dataset Statistics}

Table~\ref{tab:evasionbench_stats} presents the complete EvasionBench dataset statistics. The training data spans 2002--2022 with 8,081 unique companies, ensuring broad coverage of corporate communication patterns.

\begin{table}[t]
\centering
\small
\setlength{\tabcolsep}{4pt}
\begin{tabular}{lcccc}
\toprule
\textbf{Dataset} & \textbf{Samples} & \textbf{Comp.} & \textbf{Q len} & \textbf{A len} \\
\midrule
Train-60K & 60,000 & 6,943 & 77.5 & 149.8 \\
Train-24K & 24,000 & 4,597 & 70.0 & 136.3 \\
\textbf{Total Train} & \textbf{84,000} & \textbf{8,081} & \textbf{75.4} & \textbf{145.9} \\
\midrule
Gold-1K Eval & 1,000 & 319 & 72.0 & 131.3 \\
\bottomrule
\end{tabular}
\caption{EvasionBench dataset statistics. All splits are balanced (33.3\% per class). Comp. = unique companies; Q/A len = avg words.}
\label{tab:evasionbench_stats}
\end{table}

\section{Inter-Annotator Agreement}
\label{sec:iaa}

To validate annotation quality, a second human annotator independently labeled a balanced subset of 100 samples from the Gold 1K evaluation set.

\begin{table}[t]
\centering
\small
\begin{tabular}{lr}
\toprule
\textbf{Metric} & \textbf{Value} \\
\midrule
Total Samples & 100 \\
Agreement Count & 89 (89.0\%) \\
Cohen's Kappa & 0.835 \\
Macro-F1 & 88.99\% \\
\bottomrule
\end{tabular}
\caption{Inter-annotator agreement statistics.}
\label{tab:iaa}
\end{table}

The Cohen's Kappa of 0.835 indicates ``Almost Perfect'' agreement according to \citet{landis1977measurement}. Notably, 10 of 11 disagreements involve the \textit{intermediate} class, confirming it as the most ambiguous category.

\section{Eva-4B Model}
\label{sec:model}

\subsection{Base Model Selection}

We select Qwen3-4B-Instruct-2507 \citep{qwen3technicalreport} as our base model due to its strong instruction-following capability (IFEval: 83.4\%) and efficient parameter count enabling practical deployment.

\subsection{Two-Stage Training Pipeline}

\begin{figure}[t]
    \centering
    \includegraphics[width=\columnwidth]{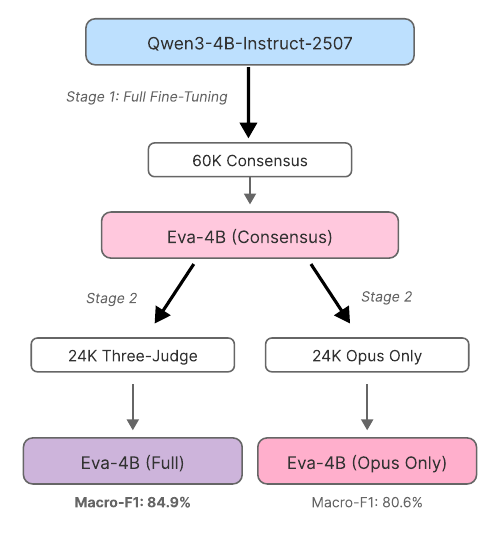}
    \caption{Two-stage training pipeline for Eva-4B. Stage 1 trains on 60K consensus data to obtain Eva-4B (Consensus). Stage 2 continues training on 24K samples with either three-judge majority labels (Full) or Opus-only labels (Opus Only).}
    \label{fig:training_pipeline}
\end{figure}

Figure~\ref{fig:training_pipeline} illustrates our two-stage training approach.

\paragraph{Stage 1: Consensus Training}
Full fine-tuning on 60K consensus samples where both frontier LLMs agreed. Training uses MS-SWIFT framework with 2 epochs, learning rate 2e-5, and bfloat16 precision.

\paragraph{Stage 2: Judge-Refined Training}
Continued fine-tuning on 24K samples including arbitrated disagreement cases. This stage incorporates the harder boundary cases resolved through three-judge voting.

\subsection{Model Variants}

We train three variants for ablation:
\begin{itemize}
    \item \textbf{Eva-4B (Consensus)}: Stage 1 only
    \item \textbf{Eva-4B (Opus Only)}: Stage 1 + Stage 2 with Opus labels
    \item \textbf{Eva-4B (Full)}: Stage 1 + Stage 2 with majority voting
\end{itemize}

\section{Experiments}
\label{sec:experiments}

\subsection{Evaluation Setup}

We evaluate 12 models on the Gold 1K evaluation set:
\begin{itemize}
    \item \textbf{Closed-source}: Claude Opus 4.5, GPT-5.2, Gemini 3 Flash
    \item \textbf{Open-source}: GLM-4.7, Qwen3-Coder, MiniMax-M2.1, Kimi-K2, DeepSeek-V3.2
    \item \textbf{Eva-4B variants}: Full, Opus Only, Consensus
    \item \textbf{Base model}: Qwen3-4B (before fine-tuning)
\end{itemize}

\subsection{Main Results}

Table~\ref{tab:main_results} presents the full evaluation results with per-class F1 scores.

\begin{table}[t]
\centering
\small
\setlength{\tabcolsep}{3pt}
\begin{tabular}{lccccc}
\toprule
\textbf{Model} & \textbf{M-F1} & \textbf{F1-D} & \textbf{F1-I} & \textbf{F1-E} \\
\midrule
Eva-4B (Full) & \textbf{84.9} & 82.2 & \textbf{80.1} & \textbf{92.4} \\
Gemini 3 Flash & 84.6 & \textbf{84.6} & 78.3 & 91.0 \\
Claude Opus 4.5 & 84.4 & 82.4 & 79.3 & 91.5 \\
GLM-4.7 & 82.9 & 84.4 & 74.7 & 89.6 \\
Eva-4B (Cons.) & 81.4 & 79.4 & 75.0 & 89.7 \\
GPT-5.2 & 80.9 & 75.4 & 76.1 & 91.2 \\
Eva-4B (Opus) & 80.6 & 77.6 & 73.9 & 90.3 \\
Qwen3-Coder & 78.2 & 72.1 & 72.4 & 90.0 \\
\midrule
Qwen3-4B (Base) & 34.3 & 7.3 & 33.3 & 62.3 \\
\bottomrule
\end{tabular}
\caption{Model performance on Gold 1K. M-F1 = Macro-F1; F1-D/I/E = F1 for Direct/Intermediate/Evasive. \textbf{Bold} = best in column. All models struggle most with Intermediate (F1-I).}
\label{tab:main_results}
\end{table}

\subsection{Ablation Study}

\begin{figure}[t]
    \centering
    \includegraphics[width=\columnwidth]{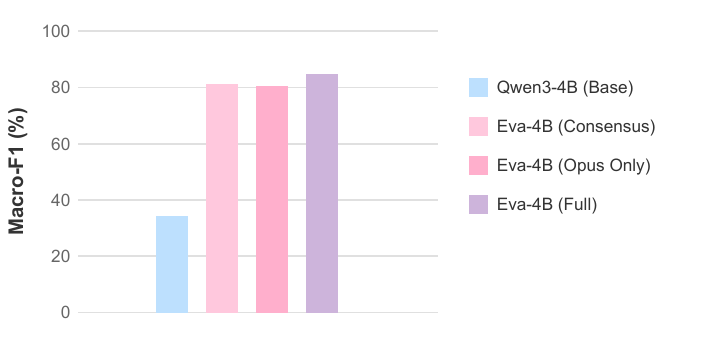}
    \caption{Ablation study: Macro-F1 comparison of Eva-4B variants. Full fine-tuning with three-judge consensus achieves 84.9\%, a +50.6 pp improvement over the base model.}
    \label{fig:ablation}
\end{figure}

Figure~\ref{fig:ablation} shows the ablation results. Our ablation demonstrates the value of multi-model consensus:

\begin{itemize}
    \item \textbf{Base $\rightarrow$ Consensus}: +47.1 pp Macro-F1
    \item \textbf{Consensus $\rightarrow$ Full}: +3.5 pp Macro-F1
    \item \textbf{Full vs. Opus Only}: +4.3 pp Macro-F1
\end{itemize}

The three-judge majority voting (Full) outperforms single-model labeling (Opus Only), validating the MMC approach.

\begin{figure}[t]
    \centering
    \includegraphics[width=\columnwidth]{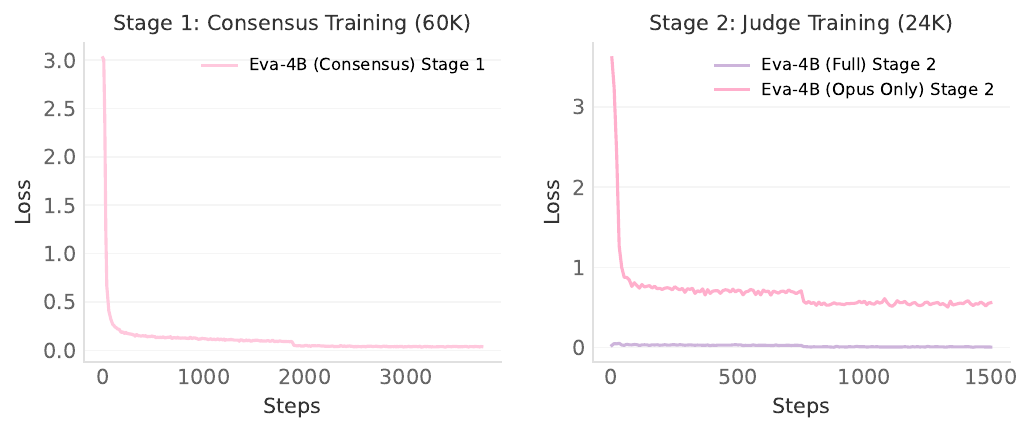}
    \caption{Training loss curves for Eva-4B variants. Stage 2 training shows dramatically different convergence: Eva-4B (Full) with three-judge labels reaches loss 0.007, while Eva-4B (Opus Only) plateaus at 0.56, indicating noisier single-model annotations.}
    \label{fig:training_loss}
\end{figure}

\paragraph{Training Dynamics}
Figure~\ref{fig:training_loss} reveals a striking difference in training convergence. Eva-4B (Full) achieves a final loss of 0.007, while Eva-4B (Opus Only) converges to 0.56---an 80$\times$ difference. This suggests that three-judge consensus labels produce more consistent training signals, while single-model labels contain noise that impedes learning. This finding provides strong evidence for the effectiveness of the MMC annotation framework.

\subsection{Error Analysis}

\begin{figure}[t]
    \centering
    \includegraphics[width=\columnwidth]{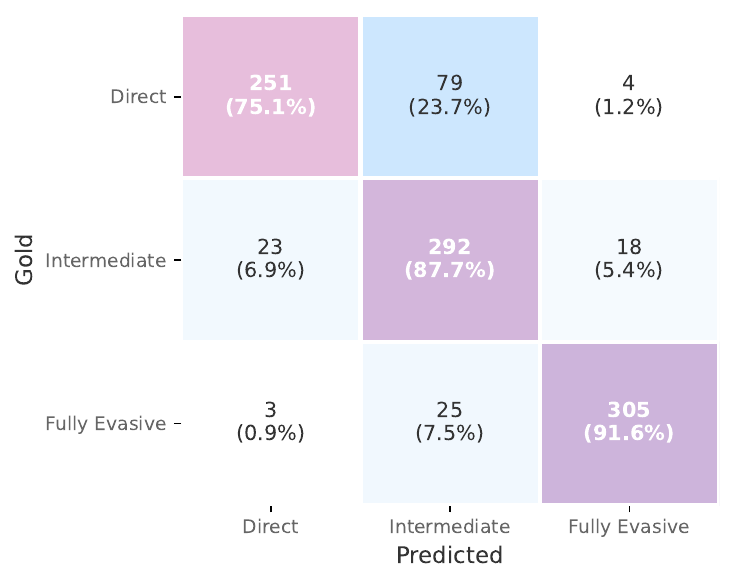}
    \caption{Confusion matrix for Eva-4B (Full) on the Gold 1K evaluation set. The model achieves 84.8\% accuracy with most errors occurring between adjacent classes.}
    \label{fig:confusion}
\end{figure}

Figure~\ref{fig:confusion} shows the confusion matrix for Eva-4B (Full). Analysis reveals:
\begin{itemize}
    \item 52.0\% of errors: direct $\rightarrow$ intermediate
    \item Adjacent class confusion accounts for 95.4\% of errors
    \item \textit{Fully evasive} is easiest to detect (F1: 92.4\%)
\end{itemize}

\section{Discussion}
\label{sec:discussion}

\paragraph{Why is Intermediate difficult?}
The intermediate class represents subtle evasion where executives appear to address questions while actually sidestepping the core ask. This ambiguity challenges both humans (IAA disagreements) and models.

\paragraph{Qualitative Error Analysis}
We manually examined Eva-4B's 152 errors on the Gold 1K set. Three patterns emerge:

\textbf{(1) Hedging-induced over-prediction of evasion.} The dominant error type (52\%) involves classifying \textit{direct} responses as \textit{intermediate}. These cases typically feature executives providing clear answers but using hedging language (``we do expect,'' ``soon'') that triggers false evasion signals. For example, when asked ``Is that still expected to occur?'' the response ``We do expect it to occur soon'' is gold-labeled \textit{direct} but predicted \textit{intermediate}---the model mistakes temporal vagueness for evasion.

\textbf{(2) Qualitative vs.\ quantitative directness.} When analysts ask quantitative questions (``how much,'' ``what percentage''), executives sometimes provide qualitative explanations listing specific actions without numbers. The model classifies these as \textit{direct} due to their specificity, while human annotators label them \textit{intermediate} for failing to address the numeric core.

\textbf{(3) Shared difficulty across models.} Only 10.5\% of Eva-4B errors are unique to our model; 33.6\% of error samples are misclassified by 5--6 of the top models, suggesting these represent genuinely ambiguous cases at the boundary of human annotation reliability.

\begin{figure}[t]
    \centering
    \includegraphics[width=0.8\columnwidth]{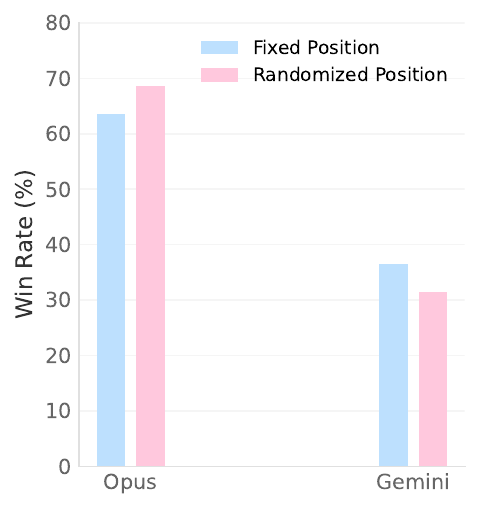}
    \caption{Position bias analysis: Fixed position (always Opus first) vs. randomized position. Randomization reveals a +5.1\% win rate shift, confirming position bias in LLM judges.}
    \label{fig:position_bias}
\end{figure}

\paragraph{Position Bias in LLM-as-Judge}
To validate our anti-bias mechanism, we conducted a controlled experiment comparing fixed-position vs. randomized-position judge assignments on 5,541 samples. Figure~\ref{fig:position_bias} shows the results: randomization increased Opus's win rate from 63.5\% to 68.6\% (+5.1\%), demonstrating substantial position bias in LLM-as-judge settings. This finding validates the necessity of randomized presentation order in our methodology.

\paragraph{Practical Applications}
EvasionBench enables: (1) automated screening for investor relations, (2) regulatory monitoring of corporate disclosure quality, (3) academic research on strategic communication.

\section{Conclusion}
\label{sec:conclusion}

We present EvasionBench, the first large-scale benchmark for detecting managerial evasion in earnings calls.
Our Multi-Model Consensus framework provides a scalable approach to annotation that outperforms single-model labeling.
Eva-4B demonstrates that a 4B parameter model can match or exceed frontier LLMs on this task when trained with high-quality consensus data.
We release all data, models, and code to facilitate future research.

\section*{Limitations}

\paragraph{Domain Specificity}
Our dataset and models are trained exclusively on earnings call transcripts. Generalization to other domains (political interviews, legal depositions) requires further validation.

\paragraph{Annotation Scale}
While our Gold 1K set is validated, single-annotator labeling with 100-sample IAA verification represents a limitation. Larger-scale human annotation would strengthen validity.

\paragraph{Temporal Coverage}
Our data spans 2002-2022. Language patterns and evasion strategies may evolve, requiring periodic updates.

\paragraph{English Only}
The current benchmark covers only English-language transcripts from U.S.-listed companies.

\section*{Ethics Statement}

This research uses publicly available earnings call transcripts. No private or sensitive personal information is included. The evasion detection system is intended for research and analytical purposes; we caution against using model outputs as sole evidence for legal or regulatory actions. The dataset will be released under appropriate academic licenses.

\bibliography{references}

\appendix

\section{Training Hyperparameters}
\label{app:hyperparams}

\begin{table}[h]
\centering
\small
\begin{tabular}{lcc}
\toprule
\textbf{Parameter} & \textbf{Stage 1} & \textbf{Stage 2} \\
\midrule
Framework & MS-SWIFT & MS-SWIFT \\
Training Type & Full & Full \\
Base Model & Qwen3-4B & Eva-4B-Cons. \\
Dataset Size & 60K & 24K \\
Epochs & 2 & 2 \\
Learning Rate & 2e-5 & 2e-5 \\
Batch Size (per GPU) & 8 & 8 \\
Gradient Accumulation & 2 & 1 \\
Effective Batch Size & 32 & 32 \\
Max Length & 2500 & 2048 \\
Precision & bfloat16 & bfloat16 \\
Warmup Ratio & 3\% & 3\% \\
\bottomrule
\end{tabular}
\caption{Training hyperparameters for two-stage fine-tuning.}
\label{tab:hyperparams}
\end{table}

\section{Full Model Results}
\label{app:full_results}

Table~\ref{tab:full_results} presents complete evaluation results for all 12 models on the Gold 1K evaluation set.

\begin{table}[h]
\centering
\small
\setlength{\tabcolsep}{3pt}
\begin{tabular}{rlcccccc}
\toprule
\textbf{\#} & \textbf{Model} & \textbf{Cat.} & \textbf{Acc} & \textbf{M-F1} & \textbf{F1-D} & \textbf{F1-I} & \textbf{F1-E} \\
\midrule
1 & Eva-4B (Full) & Eva & \textbf{84.8} & \textbf{84.9} & 82.2 & \textbf{80.1} & \textbf{92.4} \\
2 & Gemini 3 Flash & Closed & 84.6 & 84.6 & \textbf{84.6} & 78.3 & 91.0 \\
3 & Claude Opus 4.5 & Closed & 84.1 & 84.4 & 82.4 & 79.3 & 91.5 \\
4 & GLM-4.7 & Open & 83.1 & 82.9 & 84.4 & 74.7 & 89.6 \\
5 & Eva-4B (Cons.) & Eva & 81.0 & 81.4 & 79.4 & 75.0 & 89.7 \\
6 & GPT-5.2 & Closed & 80.8 & 80.9 & 75.4 & 76.1 & 91.2 \\
7 & Eva-4B (Opus) & Eva & 80.6 & 80.6 & 77.6 & 73.9 & 90.3 \\
8 & Qwen3-Coder & Open & 78.0 & 78.2 & 72.1 & 72.4 & 90.0 \\
9 & MiniMax-M2.1 & Open & 71.8 & 71.3 & 72.2 & 59.6 & 82.1 \\
10 & DeepSeek-V3.2 & Open & 66.7 & 66.9 & 61.4 & 64.3 & 75.0 \\
11 & Kimi-K2 & Open & 67.8 & 66.7 & 66.8 & 53.6 & 79.6 \\
12 & Qwen3-4B (Base) & Base & 42.3 & 34.3 & 7.3 & 33.3 & 62.3 \\
\bottomrule
\end{tabular}
\caption{Full evaluation results for all 12 models. Cat. = Category (Eva = Eva-4B variants, Closed = closed-source, Open = open-source, Base = base model). All metrics in \%. \textbf{Bold} = best in column.}
\label{tab:full_results}
\end{table}

\end{document}